\documentclass[runningheads]{llncs}
\usepackage{graphicx}
\usepackage{amsmath,amssymb} %
\usepackage{color}
\usepackage{setspace}

\begin{document}
\pagestyle{headings}
\mainmatter

\title{Deep attention-based classification network for robust depth prediction} %
\titlerunning{Attention-based classification networks for depth prediction}
\authorrunning{Li \it et al.}

\author{Ruibo Li$^{1\star}$, Ke Xian$^{1}$\thanks{The first two
authors contributed equally.},
Chunhua Shen$^2$, Zhiguo Cao$^1$, Hao Lu$^1$, Lingxiao Hang$^1$}
\institute{$^1$Huazhong University of Science and Technology, China\\
	$^2$The University of Adelaide, Australia\\
	{e-mail: \small \tt \{liruibo, kexian\}@hust.edu.cn}
}

\maketitle

\begin{abstract}
	In this paper, we present our deep attention-based classification (DABC) network for robust single image depth prediction, 
	in the context of the Robust Vision Challenge 2018 (ROB 2018)\footnote{\footnotesize\texttt{http://www.robustvision.net/index.php}}. Unlike conventional depth prediction,
	our goal is to design a model that can perform well in both indoor and outdoor scenes with a single parameter set. However, robust depth prediction suffers from two 
	challenging problems: 
	a) How to extract more discriminative features for different scenes (compared to a single scene)?
	b) How to handle the large differences of depth ranges between indoor and outdoor datasets?
	To address these two problems, we first formulate depth prediction as a multi-class classification task and apply a softmax classifier to classify the depth label of each pixel. We then introduce a global pooling layer and a channel-wise attention mechanism to 
	adaptively select the discriminative channels of features and to update the original features by assigning important channels with higher weights.
	Further, to reduce the influence of quantization errors, we employ a soft-weighted sum inference strategy 
	for the final prediction.

	Experimental results on both indoor and outdoor datasets demonstrate the effectiveness of our method. 
	It is worth mentioning that we won the 2-nd place in single image depth prediction entry of ROB 2018, in conjunction with IEEE Conference on Computer Vision and Pattern Recognition (CVPR) 2018.
\end{abstract}

\newpage

\section{Introduction}	
\begin{figure*}
	\begin{center}
		\includegraphics[width=0.9\linewidth]{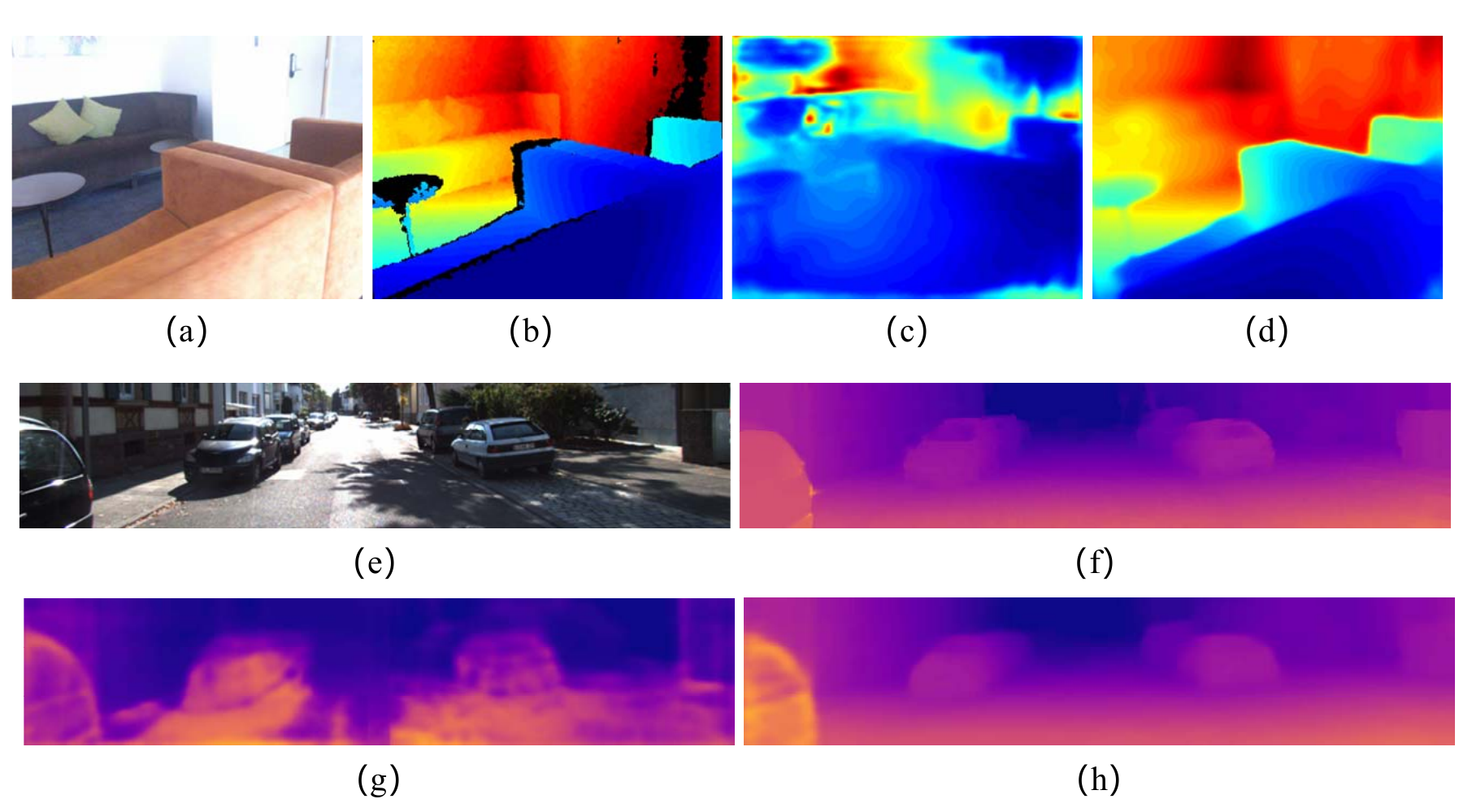}
	\end{center}
	\caption{Some examples on robust depth prediction. (a) and (e) are the input images from ScanNet and KITTI, respectively. (b) and (f) are corresponding ground truth depth maps. (c) is a depth map predicted by a model trained on KITTI. (g) is a depth map predicted by another model trained on ScanNet. (d) and (h) are the results predicted by our DABC model.}
	\label{fig_problem}
\end{figure*}

Single image depth prediction is a challenging task that aims to recover pixel-wise depth from monocular images, which plays an important role in many applications, such as 3D modeling, autonomous driving, and 2D-to-3D conversion. 
With the prosperity of deep convolutional neural networks (CNNs), many deep learning-based methods~\cite{eigen2014depth,liu2016learning,xian2018monocular} have achieved state-of-the-art performance on various RGB-D datasets, such as NYUv2~\cite{silberman2012indoor}, ScanNet~\cite{dai2017scannet}, and KITTI~\cite{geiger2013vision}. 
However, all these methods are trained individually on each dataset, which makes models be specific to certain domains.
For example, ScanNet only focuses on indoor scenes, while KITTI considers outdoor views.
The large differences between indoor and outdoor patterns limit the generalization ability of the model. That is, a model achieves remarkable performance on one dataset but performs poorly on the other one, as shown in Fig.~\ref{fig_problem}. 
In this paper, we study how to make the model robust so that it can perform well on diverse datasets.

One challenge for robust depth prediction is how to extract discriminative features from diverse scenes. Compared with conventional depth prediction, robust depth prediction requires one single model to perform well in both indoor and outdoor scenes. Thus, the model should adapt to diverse scene layouts and can tackle complex depth patterns.
Another challenge is the large difference of depth ranges between indoor and outdoor.
For example, the depth range of ScanNet is from 0$m$ to 10$m$, while the depth range of KITTI is between 2.5$m$ and 80$m$. It is worth noting that two datasets exist an overlapped depth range from 2.5$m$ to 10$m$.
In this range, different layouts belonging to indoor or outdoor scenes may correspond to the same depth value, which increases the difficulty of training models.    

In this paper, we present a deep attention-based classification network (DABC) to address these problems.
In order to tackle diverse scenes, our model learns a universal RGB-to-depth mapping from both the ScanNet and KITTI datasets. Our model is based on a U-shape~\cite{xian2018monocular,ronneberger2015u,lin2016refinenet} structure, where skip-connections fuse multi-scale depth cues to generate accurate and high-resolution results.
To extract discriminative features, we employ a global average pooling layer and a channel-wise attention mechanism to our model.
Through the global average pooling layer, the model can capture the channel-wise statistics of features that represent the global information and visual characteristics of a scene.
According to the channel-wise statistics, the attention mechanism assigns important channels with higher weights and updates the original features. 
In addition, to tolerate the significant difference of depth ranges, we formulate depth prediction as a multi-class classification problem.
By discretizing continuous depth values into several intervals, we choose a softmax layer as the predictor. 
In this way, each neuron in the last layer only needs to activate for a specific depth interval rather than to predict the depth value in the whole depth range, which makes the model easy to train.
Further, to reduce the influence of quantization error caused by depth discretization, we employ a soft-weighted sum inference method~\cite{boli2018} to obtain the final continuous predictions.

The contributions of this work can be summarized as follows:

$\bullet$ We present one of the first explorations on the task of robust depth prediction 
and propose a deep attention-based classification network (DABC).

$\bullet$ For the task of depth prediction, we are the first to employ the channel-wise attention mechanism, and verify the effectiveness of the attention module on choosing scene-specific features.

$\bullet$ Our method achieves state-of-the-art performance on both the ScanNet and KITTI datasets, and won the 2-nd place in the single image depth prediction entry of ROB 2018.

\section{Related Work}	
\textbf{Metric depth prediction.}
Recently, many methods  have achieved impressive performance by employing CNNs to monocular depth prediction.
In order to regress the depth map from a RGB image, Eigen et al.~\cite{eigen2014depth} proposed a multi-scale architecture which consists of two CNNs: the first one predicts the coarse depth map based on global context, the second one refines the coarse prediction using local information. 
Laina et al.~\cite{laina2016deeper} applied the residual learning~\cite{he2016deep} to the depth prediction task and presented an encoder-decoder model with a novel up-sampling module.
In order to encourage spatial consistency in the predicted depth map, some researchers tried to formulate a structured regression problem by combining CNNs with Conditional Random Fields (CRFs).
For instance, Liu et al.~\cite{liu2016learning} proposed an end-to-end network by incorporating a CRF loss layer into a generic superpixel-wise CNN. The spatial relation of adjacent superpixels can be modeled by a CRF, and the parameters of CNN and CRF can be jointly optimized.
Instead of just refining the coarse predictions, Xu et al.~\cite{xu2017multi} designed a novel continuous CRF framework to merge multi-scale features derived from CNN inner layers.

Aside from regressing the exact depth values from a RGB image, 
Cao et al.~\cite{cao2017estimating} recast depth estimation as a pixel-wise classification problem to classify the depth range for each pixel.
Further, Li et al.~\cite{boli2018} proposed a soft-weighted-sum inference strategy to replace the hard-max inference used in~\cite{cao2017estimating} to obtain continuous depth predictions.
However, the above models are only trained on indoor and outdoor datasets individually, which greatly limits the application of these models.
In contrast, we propose an attention-based classification network which is adaptive for a variety of scenes and achieves the state-of-the-art performance on the ScanNet and KITTI datasets with a single parameter set.

\noindent \textbf{Relative depth prediction.}
Some recent works~\cite{xian2018monocular,chen2016single} focus on recovering depth ordering from unconstrained monocular  images instead of predicting the metric depth value directly.
The idea behind relative depth prediction is similar to robust depth prediction, because they all need models to be applicable to diverse unconstrained natural scenes.
However, when using a model trained on relative depth data to predict metric depth, a transformation function that converts relative depth values to metric depth values should be provided for each dataset separately, which reduces the practicality of the model.
Thus, a model that can directly estimate metric depth from unconstrained images is more practical and effective.  

\noindent \textbf{Attention mechanism.}
Recently, attention mechanism has been successfully applied to various tasks, such as image classification~\cite{hu2017squeeze,wang2017residual}, monocular depth prediction~\cite{xu2018structured,kong2018pixel}, and semantic segmentation~\cite{chen2016attention,yu2018learning}.
Kong et al.~\cite{kong2018pixel} presented a pixel-wise attentional gating unit to learn spatial allocation of computation in dense labeling tasks.
Xu et al.~\cite{xu2018structured} proposed an attention-based CRF where a structured attention model~\cite{kim2017structured} is applied to regulate the feature fusion across different scales.
Unlike above methods, we focus on how to extract more discriminative features for diverse scenes with a channel-wise attention mechanism.

\section{Method}	
In order to use a single model to estimate the depth map from an unconstrained image, we present a deep attention-based classification network (DABC) for robust depth prediction.
As illustrated in Fig.~\ref{fig_architecture}, we reformulate robust depth prediction as a classification task and learn a universal RGB-to-depth mapping that is suitable for both indoor and outdoor scenes. In the following, we first describe the architecture of our DABC model. Then we discuss our discretization strategy and explain the technical details of learning and inference stages.

\subsection{Network Architecture}
\begin{figure*}
	\begin{center}
		\includegraphics[width=1\linewidth]{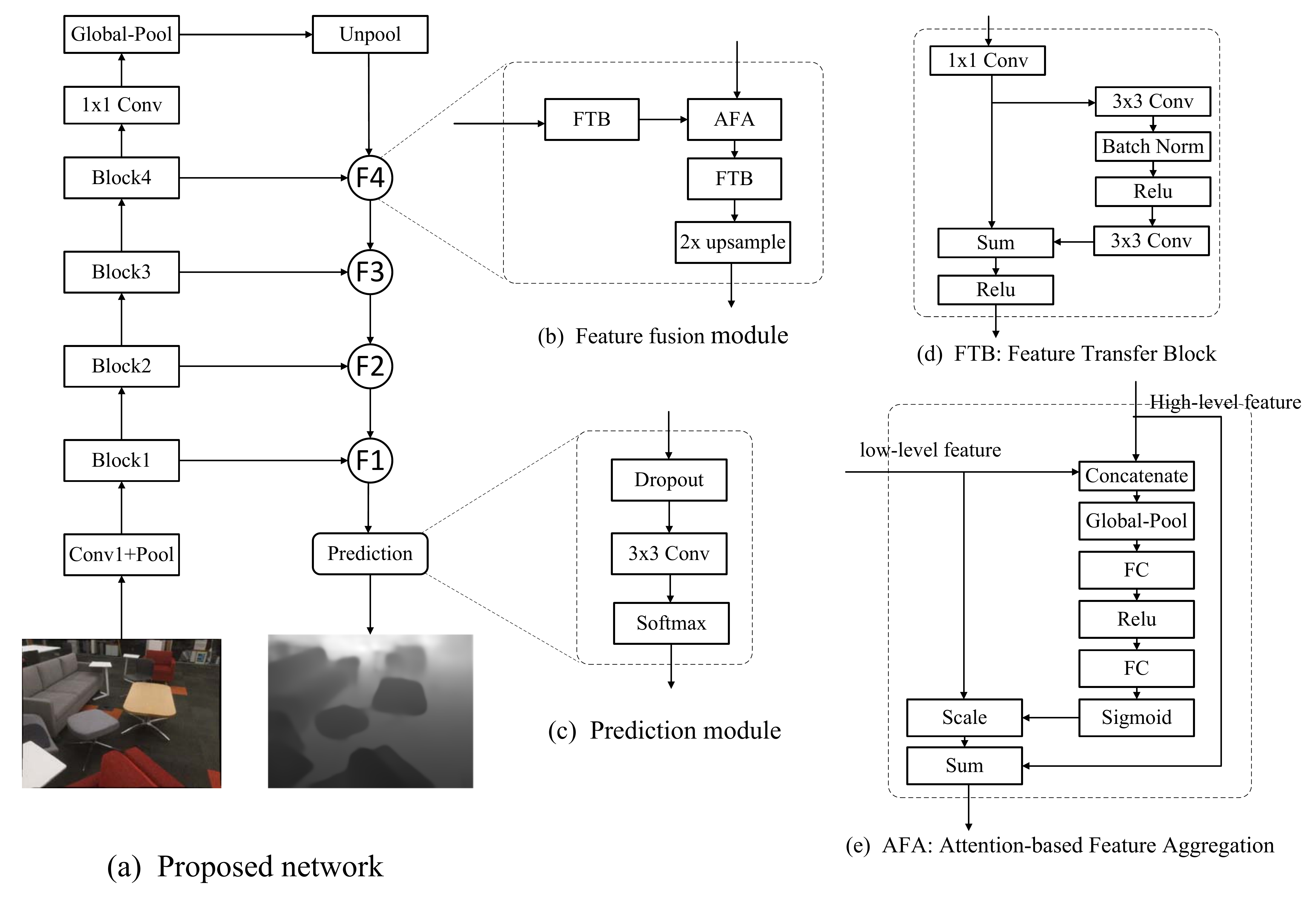}
	\end{center}
	\caption{\textbf{Illustration of our DABC model for robust depth prediction.} (a) is our proposed network. (b) shows the process of multi-scale feature fusion. (c) is a prediction module to generate score maps. (d) is a feature transfer block. (e) is an attention-based feature aggregation block to fuse multi-scale features.} 
	\label{fig_architecture}
\end{figure*}

As shown in Fig.~\ref{fig_architecture}, our deep attention-based classification model consists of three parts: feature extraction module, multi-scale feature fusion module, and prediction module.
The feature extraction module is built upon ResNeXt-101 (64$\times$4d)~\cite{xie2017aggregated}.  
To make it suitable for the task of depth prediction, we first remove the last pooling layer, fully-connected layer and softmax layer. We divide ResNeXt into 4 blocks according to the size of the output feature maps. To capture global information, we attach a 1$\times$1 convolution layer, global average pooling layer and unpooling layer to the top of the modified ResNeXt. Specifically, the 1$\times$1 convolution layer is used to reduce the dimension of the feature maps from block4, and the global pooling layer is used to encode the global context features. The unpooling layer is used to restore the spatial resolution of the global features. 

Due to max pooling and convolutional strides in ResNeXt, the size of final feature map is 1/32 of the original input image.
Inspired by~\cite{xian2018monocular,yu2018learning,lin2017refinenet}, we use a progressive refinement strategy, which aggregates high-level semantic features and low-level fine-grained features layer-by-layer, to generate high-resolution predictions.
In each feature fusion block, we first use a feature transfer block (FTB)~\cite{yu2018learning} to transfer the features from ResNeXt suitable for the task of depth prediction. 
The 1$\times$1 Convolution layer in FTB is used to unify the dimension of features. 
Then, two groups of features with different information are merged via an attention-based feature aggregation block (AFA)~\cite{yu2018learning}.
At the beginning of each AFA block,  two groups of features are concatenated to produce a channel-wise attention vector that represents the importance of each channel in low-level features. 
Based on the attention vector, the low-level features are selectively summed to the high-level features to generate final outputs at the current scale.  
Thus, the channel-wise attention mechanism is encompassed by the AFA block.
After the AFA, another FTB and a bilinear upsampling operation are stacked to enlarge the spatial resolution of the fused features. 
More specifically, the channel number of AFA connected to Block4 is 512 and the others are 256. 
In order to produce final score maps, we attach a prediction module to the end of the fusion module, which consists of a dropout layer, a 3$\times$3 convolution layer and a softmax-loss layer.

\subsection{Depth Discretization Strategy}

To formulate depth prediction as a classification task, we discretize continuous depth values into several discrete bins in the log space. 
Assuming that the depth range is  $\left[ {\alpha ,\beta } \right]$, the discretization strategy can be defined as
\begin{equation}
	\begin{array}{l}
		l = round\left( {{{\left( {{{\log }_{10}}\left( d \right) - {{\log }_{10}}\left( \alpha  \right)} \right)} \mathord{\left/
					{\vphantom {{\left( {{{\log }_{10}}\left( d \right) - {{\log }_{10}}\left( \alpha  \right)} \right)} q}} \right.
					\kern-\nulldelimiterspace} q}} \right),\\
		q = {{\left( {{{\log }_{10}}\left( \beta  \right) - {{\log }_{10}}\left( \alpha  \right)} \right)} \mathord{\left/
				{\vphantom {{\left( {{{\log }_{10}}\left( \beta  \right) - {{\log }_{10}}\left( \alpha  \right)} \right)} {K,}}} \right.
				\kern-\nulldelimiterspace} {K,}}
	\end{array}
\end{equation}
where $l$ is the quantized label, $d$ is the original continuous depth value, $K$ is the number of sub-intervals and $q$ is the width of the quantization bin.   
Specifically, $\left[ {\alpha ,\beta } \right]$ and $K$ are set to $\left[ {0.25 ,80 } \right]$ and 150, respectively.

\subsection{Learning and Inference}

At the training stage, we use the pixel-wise multinomial logistic loss function to train our DABC network:
\begin{equation}
	L =  - \frac{1}{N}\sum\limits_{i = 1}^N {\sum\limits_{D = 1}^K {1\left\{ {{y^{\left( i \right)}} = D} \right\}\log \left( {P\left( {D|{c_i}} \right)} \right)} } ,
\end{equation}
where $N$ is the number of valid training pixels in a mini-batch and  ${y^{\left( i \right)}}$ is the ground truth label of pixel $i$.
$P\left( {D|{c_i}} \right) = {{{e^{{c_{i,D}}}}} \mathord{\left/
		{\vphantom {{{e^{{c_{i,D}}}}} {\sum\nolimits_d^K {{e^{{c_{i,{\rm{d}}}}}}} }}} \right.
		\kern-\nulldelimiterspace} {\sum\nolimits_d^K {{e^{{c_{i,{\rm{d}}}}}}} }}$
represents the probability of pixel $i$ labelled with $D$, and ${{c_{i,{\rm{d}}}}}$ is the output from the last 3$\times$3 convolution layer. 

At the inference stage, we use a soft-weighted sum inference strategy~\cite{boli2018} to compute final predictions after obtaining the score maps from the softmax layer. For clarity, we use $\bf{p}_i$ to denote the score vector of pixel $i$. The depth value of pixel $i$ can be calculated as:
\begin{equation}
	{\hat d_i} = {10^{{{\bf{w}}^{T}}{{\bf{p}}_{\bf{i}}}}},{w_j} = {\log _{10}}\left( \alpha  \right) + q*j,
\end{equation}
where ${\hat d}_i$ is the predicted depth value of pixel $i$,  $\bf{w}$ is the weight vector of bins, and $w_j$ is $j$-th element of  $\bf{w}$.

\section{Experiments}
In this section, we evaluate our DABC model on two publicly available RGB-D datasets: ScanNet~\cite{dai2017scannet} and KITTI~\cite{Uhrig2017THREEDV}.  
We first introduce the two datasets and the evaluation metrics in this challenge.
Then we give the details about our experimental setting.
Finally, we compare our DABC model with the official baseline method and other outstanding methods that participate in this challenge. 

 \subsection{Datasets and Metrics}
\textbf{Datasets.}
The ScanNet~\cite{dai2017scannet} is a large-scale RGB-D dataset for indoor scene reconstruction, which contains 2.5M RGB-D images in 1513 scenes.
Original RGB images are captured at a resolution of 1296$\times$968 and depth at 640$\times$480 pixels.
Specially, the resized RGB images of the resolution 640$\times$480 are also officially provided for the robust depth prediction task.  

The KITTI~\cite{Uhrig2017THREEDV} contains street scene RGB-D videos captured by cameras and a LiDAR scanner.
The resolution of RGB-D pairs is about 376$\times$1242.
In the KITTI dataset, less than $ 5{\rm{\% }}$ pixels of the raw depth map is valid, thus we fill in the missing pixels to produce dense ground truth depth maps for training via the ``colorization'' algorithm~\cite{levin2004colorization}.

\noindent \textbf{Evaluation Metrics.}
In order to measure the performance of our model, we introduce several evaluation metrics used in this robust vision challenge:

$\bullet$\quad The mean relative absolute error (absRel):\quad$\frac{1}{Q}\sum\nolimits_{i = 1}^Q {\frac{{\left| {{d_i} - \widehat {{d_i}}} \right|}}{{{d_i}}}} $

$\bullet$\quad The mean relative squared error (sqRel):\quad$\frac{1}{Q}\sum\nolimits_{i = 1}^Q {\frac{{{{\left( {{d_i} - \widehat {{d_i}}} \right)}^2}}}{{d_i^2}}} $

$\bullet$\quad The mean absolute error of the inverse depth (imae):\quad$\frac{1}{Q}\sum\nolimits_{i = 1}^Q {\left| {{p_i} - \widehat {{p_i}}} \right|} $

$\bullet$\quad The root mean squared error of the inverse depth (irmse):

\quad$\sqrt {\frac{1}{Q}\sum\nolimits_{i = 1}^Q {\left( {{p_i} - \widehat {{p_i}}} \right)} }$

$\bullet$\quad Scale invariant error (SI):

\quad $\frac{1}{Q}\sum\nolimits_{i = 1}^Q {{{\left( {{d_i} - \widehat {{d_i}} + \alpha \left( {d,\widehat d} \right)} \right)}^2}}, $

\quad $\alpha \left( {d,\widehat d} \right) = \frac{1}{{\left| {{S_i}} \right|}}\sum\limits_{j \in {S_i}} {\left( {\widehat {{d_j}} - {d_j}} \right)} $

$\bullet$\quad Scale invariant logarithmic error (SILog):

\quad $\frac{1}{Q}\sum\nolimits_{i = 1}^Q {{{\left( {\log {d_i} - \log \widehat {{d_i}} + \alpha '\left( {d,\widehat d} \right)} \right)}^2}}, $

\quad $\alpha '\left( {d,\widehat d} \right) = \frac{1}{{\left| {{S_i}} \right|}}\sum\limits_{j \in {S_i}} {\left( {\log \widehat {{d_j}} - \log {d_j}} \right)} $

\noindent where ${{d_i}}$ and ${\widehat {{d_i}}}$ denote the ground truth and estimated depth for pixel $i$, $Q$ is the total number of valid pixels in the test images, ${{p_i}}$ and ${\widehat {{p_i}}}$ represent the inverse of ${{d_i}}$ and ${\widehat {{d_i}}}$, respectively.
${{S_i}}$ is a set of pixels that belong to the same image with pixel $i$.

\subsection{Experimental Setting}

We implement our model using the public available \mbox{\texttt{MatConvNet}}~\cite{vedaldi15matconvnet} Library on a single Nvidia GTX 1080 GPU. The feature extraction module is initialized using ResNeXt-101 (64$\times$4d), and the others are initialized with simple random Gaussian initialization. Data augmentation is performed on-the-fly during training. Specifically, horizontal flipping and randomly scaling are applied, and the default input resolution of our model is 256$\times$320. We train our model using SGD with a mini-batch size of 8. The momentum and weight decay are set to 0.9 and 0.0005, respectively. We train our model with an initial learning rate of 0.001 for 30 epochs, and divide it by 10 for another 20 epochs.

During training, for ScanNet, we first downsample images to 240$\times$320, and then pad the downsampled images to 256$\times$320 with zeros. For KITTI, we first downsample images to 182$\times$612 to make the spatial resolution of images consistent with that of the ScanNet. We then randomly crop the images to the size of 182$\times$320 by keeping the height constant, followed by padding the cropped images to 256$\times$320 with zeros. To train our model, we use 37K officially provided training data. More specifically, the number of images from KITTI and ScanNet are 17K and 20K, respectively.

At the testing phase, we first downsample the input image to half. If the width of the image is larger than 320, we split it to two parts by keeping the height constant and the width of 320. Then, the image or image slices are padded with zeros to form the inputs of our model. After inference, we discard the invalid regions of the output depth maps and perform a bilinear interpolation to get our final predictions. Noted that, for image slices, the depth values of overlapped regions are obtained by averaging the predictions of depth slices.

\subsection{Experiment Results}
\begin{figure*}
	\begin{center}
		\includegraphics[scale=.61]{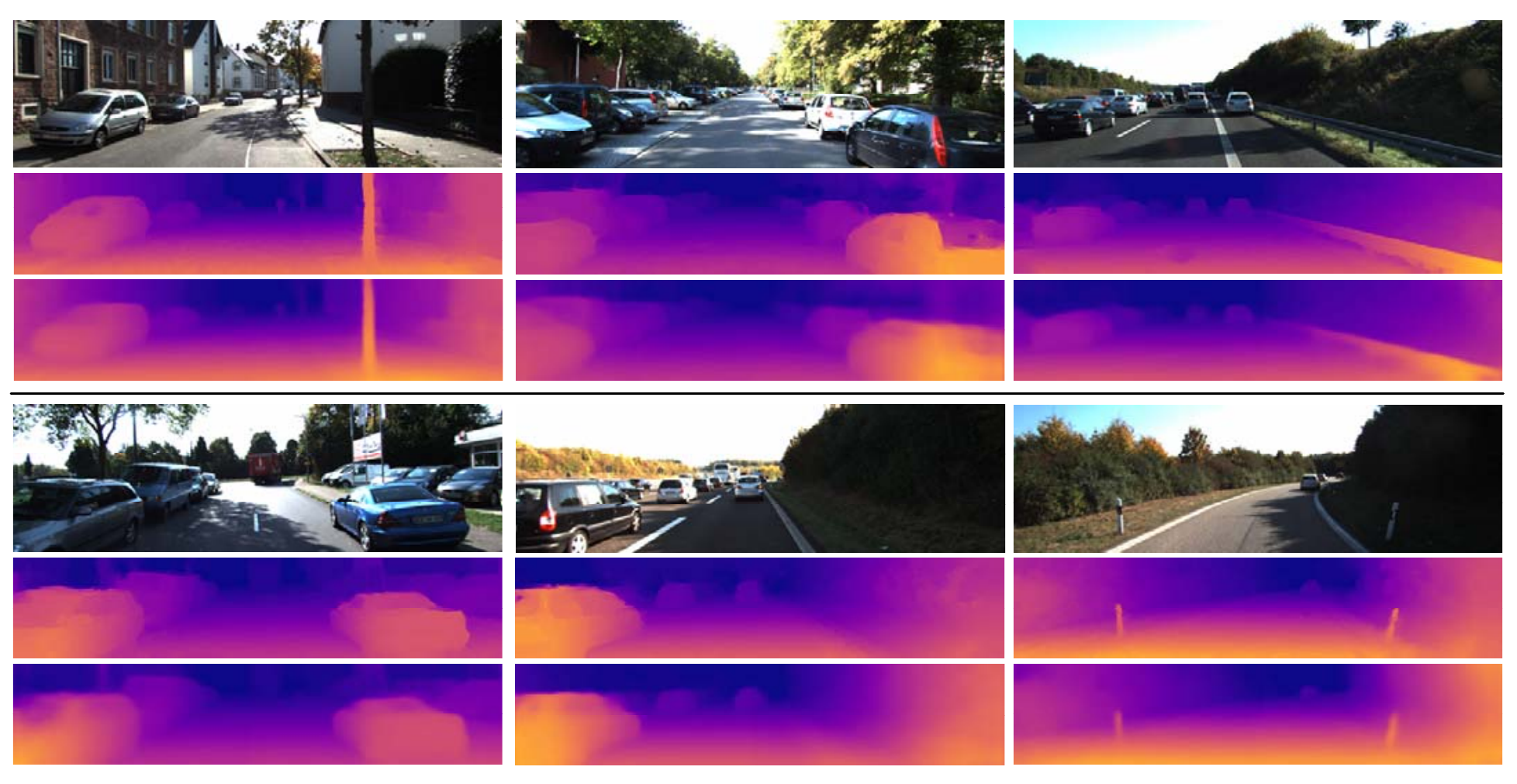}
	\end{center}
	\caption{Qualitative results of our DABC model on the KITTI validation dataset: original color images (top row), ground truth depth maps (center), and  predicted depth maps (bottom row). Color indicates depth (purple is far, yellow is close).  } 
	\label{fig_kitti}
\end{figure*}

\begin{figure*}
	\begin{center}
		\includegraphics[scale=.61]{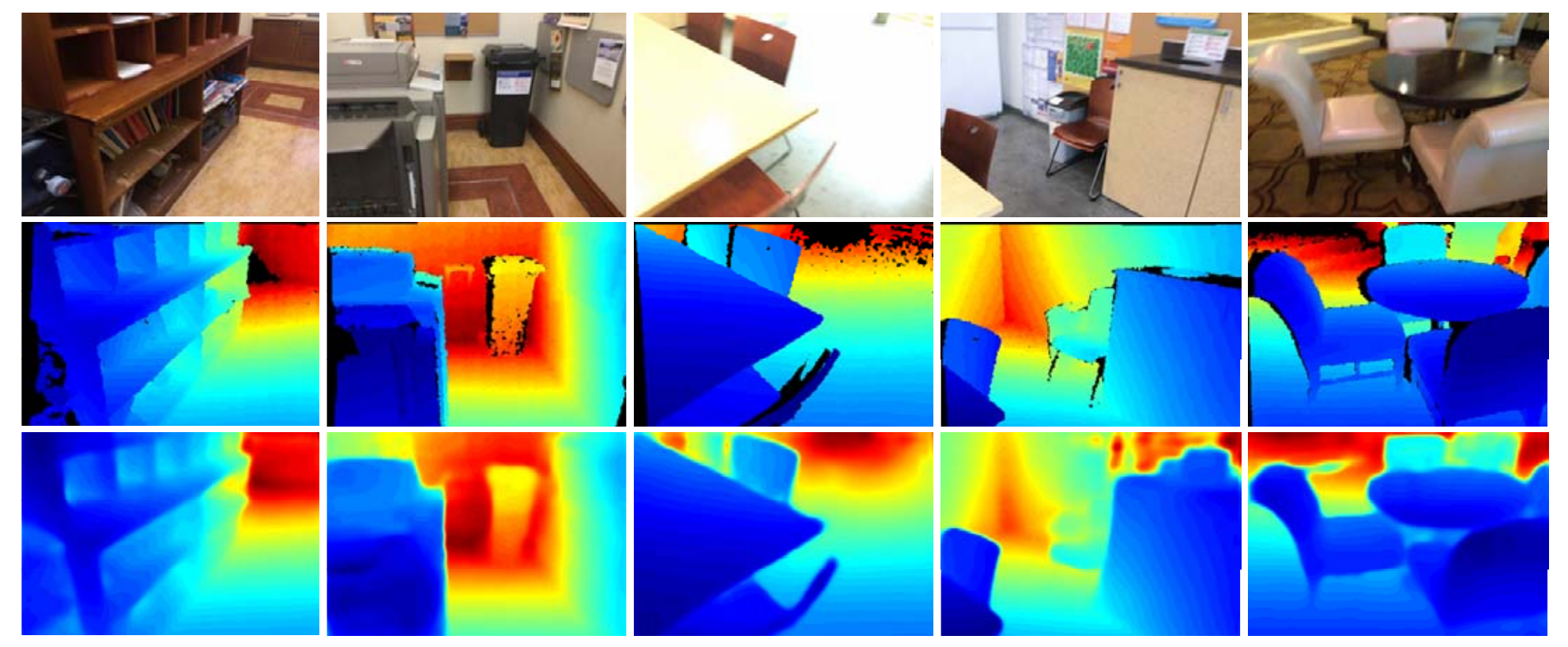}
	\end{center}
	\caption{Qualitative results of our DABC model on the ScanNet validation dataset: original color images (top row), ground truth depth maps (center), and predicted depth maps (bottom row). Color indicates depth (red is far, blue is close). The invalid regions in ground truth depth maps are painted black.} 
	\label{fig_scannet}
\end{figure*}

In this section, our DABC model is compared with the official baseline method and the other three outstanding methods that participate in this challenge. 
Table~\ref{table_KITTI} and Table~\ref{table_Scannet} show the results of the five models: 
1) \emph{Baseline}: an official baseline model based on GoogLeNetV1~\cite{szegedy2015going};
2) \emph{DORN}: a deep ordinal regression network proposed by Fu et al.~\cite{fu2018deep};
3) \emph{CSWS}: a classification model with soft-weighted-sum inference strategy trained by Li et al.~\cite{boli2018}; 
4) \emph{APMOE}: a CNN model with pixel-wise attentional gating units proposed by Kong et al.~\cite{kong2018pixel};
5) \emph{DABC}: our proposed deep attention-based classification network. 
As shown in Table~\ref{table_KITTI} and Table~\ref{table_Scannet}, the errors of our DABC model are significantly lower than that of the official baseline model about by $40\%  \sim 70\% $ on both the KITTI and ScanNet.
Compared with other methods, our DABC model achieves the first place on the ScanNet and the third place on the KITTI.
Finally, our DABC model won the 2-nd place in single image depth prediction of ROB 2018.
Some qualitative results are shown in Figure~\ref{fig_kitti} and Figure~\ref{fig_scannet}, we observe our model well captures local image details and
yields predictions of high visual quality in both outdoor and indoor scenes.

\section{Discussion}
In this section, we conduct a in-depth analysis of our DACB model.
In following experiments, we use the ResNext-50 (32x4d) to replace the original ResNext-101 (64x4d) in our DABC model for fast training. Unless otherwise stated, the models are trained with the mixed ScanNet and KITTI data. First, we compare the DABC model against a conventional regression network. Second, we evaluate the effectiveness of the attention module in choosing discriminative features.
\begin{table*}
	\vspace{1pt}
	\caption{ Performance on the KITTI benchmark.}
	\label{table_KITTI}
	\renewcommand\arraystretch{1.0}	
	\centering	
	\begin{tabular}{l@{\hskip 2cm}c@{\hskip 0.60cm}c@{\hskip 0.60cm}c@{\hskip 0.60cm}c}
		\hline
		{Method} & {SILog} & {sqRel} & {absRel} & { irmse} \\ \hline
		\hline
		Baseline  & 33.49 & 24.06  & 29.72  & 35.22 \\
		CSWS~\cite{boli2018}  & 14.85 & 3.48  & 11.84  & 16.38 \\
		APMOE~\cite{kong2018pixel}  & 14.74 & 3.88  & 11.74  & 15.63 \\		
		DORN~\cite{fu2018deep}  & \bf{13.53} & \bf{3.06}  & \bf{10.35}  & 15.96 \\
		DABC &  14.49 & 4.08  & 12.72  & \bf{15.53}  \\	
		\hline		
	\end{tabular}
\end{table*}

\begin{table*}
	\vspace{1pt}
	\caption{ Performance on the ScanNet benchmark.}
	\label{table_Scannet}
	\renewcommand\arraystretch{1.0}	
	\centering	
	\begin{tabular}{l@{\hskip 2cm}c@{\hskip 0.60cm}c@{\hskip 0.60cm}c@{\hskip 0.60cm}c@{\hskip 0.60cm}c}
		\hline
		{Method} & {SI} & {sqRel} & {absRel} & { irmse} &{imae}\\ \hline
		\hline
		Baseline  & 0.05 & 0.14 & 0.25  & 0.21 & 0.17\\
		APMOE~\cite{kong2018pixel}  & 0.03 & 0.10  & 0.20  & 0.16 & 0.13 \\
		CSWS~\cite{boli2018}  & \bf{0.02} & \bf{0.06}  & 0.15  & \bf{0.13} & \bf{0.10} \\		
		DORN~\cite{fu2018deep}  & \bf{0.02} & \bf{0.06}  & \bf{0.14}  & \bf{0.13} & \bf{0.10} \\
		DABC & \bf{0.02} & \bf{0.06}  & \bf{0.14}  & \bf{0.13} & \bf{0.10}  \\	
		\hline		
	\end{tabular}
\end{table*}
\subsection{Effect of multi-class classification}
In order to verify the effectiveness of multi-class classification, we compare our model against a conventional regression network.
In this experiment, the regression network is built upon the same basic structure as our DABC model and trained with a typical $\ell_2$ loss function.  
Note that the attention module of the original structure is removed in the regression network, because we find this module hinders the convergence in practice.
The experimental results are shown in Tables~\ref{table_Scannet_class_model} and~\ref{table_KITTI_class_model}.
Specially, the all metrics in Table~\ref{table_Scannet_class_model} are multiplied by 100 for easy comparison. 
In particular, ``Classification'' refers to our DABC model, ``Regression'' refers to the regression model, and the symbol ``*'' indicates that the marked model is trained and evaluated on the same dataset.

From Tables~\ref{table_Scannet_class_model} and~\ref{table_KITTI_class_model}, we observe that our DABC model is significantly better than the regression model on both the KITTI and ScanNet datasets.
By comparing ``Regression'' against ``Regression*'', we find that training on the mixed data harms the accuracy of the regression model on each individual dataset.
However, according to the results of ``Classification'' and ``Classification*'',  we do not observe performance degradation, which suggests our DABC model can better adapt to the mixed data and avoid the mutual interference of different datasets.

\begin{figure*}
	\begin{center}
		\includegraphics[scale=.6]{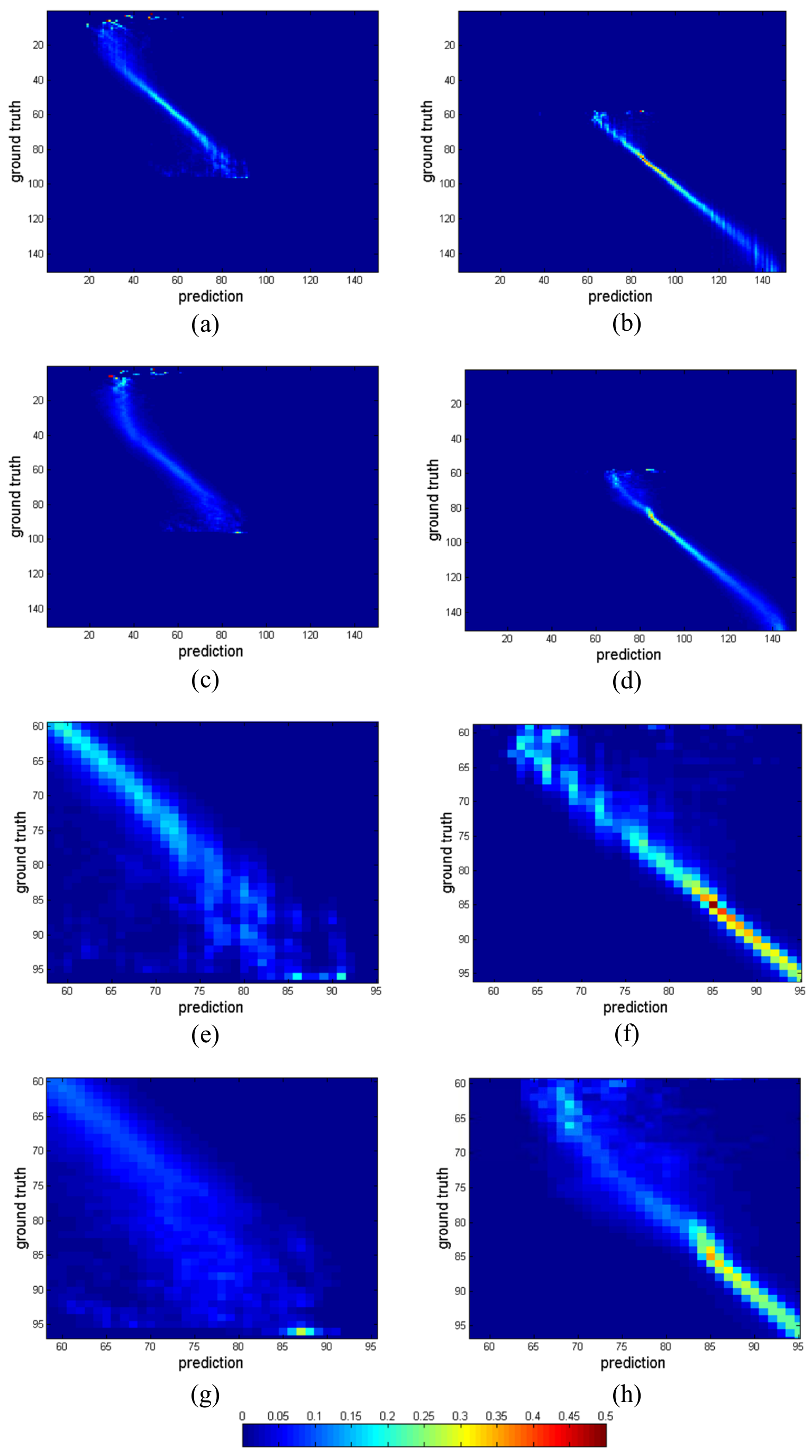}
	\end{center}
	\caption{ Visualization of confusion matrices. (a) is the confusion matrix of our DABC model on the ScanNet. (b) is the confusion matrix of our DABC model on the KITTI. (c) is the confusion matrix of the regression network on the ScanNet. (d) is the confusion matrix of the regression network on the KITTI. 	(e)$\sim$(h) show the details of the above four confusion matrices in the label range of 60 to 95, respectively. } 
	\label{fig_confusion}
\end{figure*}

Further, we draw the confusion matrices of our DABC model and the regression model on both the KITTI and ScanNet datasets.
Here, each element $\left( {i,j} \right)$ in the confusion matrix represents the empirical probability of each pixel known to be label $i$ but estimated to be label $j$.
In Figure~\ref{fig_confusion}, we can observe a clear diagonal pattern in the matrices of our DABC model, which means that the accuracy of our DABC is higher than that of the regression model on the two datasets. 
Specially, we highlight the performance of each model on the overlapped depth ranges of two datasets, i.e. $i \in \left[ {60,95} \right]$.  
From Figure~\ref{fig_confusion}(e)$ \sim $(h), we find that our DABC model performs better than the regression model on this range.
It indicates that our DABC model can tackle the complex depth patterns well in these easily-confusing depth ranges.

We believe that the outstanding performance of DABC model in robust depth prediction benefits from recasting depth prediction into  a multi-class classification problem.
When facing the complex depth patterns from indoor and outdoor scenes, in regression network, the last neuron should produce accurate activations for the whole depth patterns; but in the classification  network, each neuron in last layer only corresponds to a specified depth interval and activates for some special depth patterns, which makes the classification  network easy to train and to achieve better performance.
Specially, in the overlapped depth range, the classification network shows stronger robustness.
In practice, for each neuron of the last layer, if the depth value of a sample is far from the specified depth interval of this neuron, this sample is usually a easy negative example with a negligible loss contribution to the training of this neuron.
Thus, the samples from outdoor scenes can hardly influence the training of the neurons that correspond to the indoor depth range, and vice versa.
This is why our model can adapt to the mixed data and avoid the mutual interference of different datasets.

\begin{table*}
	\vspace{1pt}
	\caption{ Comparison between classification and regression on the ScanNet validation dataset.}
	\label{table_Scannet_class_model}
	\renewcommand\arraystretch{1.0}	
	\centering	
	\begin{tabular}{l@{\hskip 0.8cm}c@{\hskip 0.60cm}c@{\hskip 0.60cm}c@{\hskip 0.60cm}c}
		\hline
		{Method} & {sqRel} & {absRel} & { irmse} &{imae}\\ \hline
		\hline
		Regression*  & 5.93 & 17.05 & 18.65 & 11.84  \\
		Regression  & 6.51 & 17.47 & 18.65 & 11.73  \\		
		\hline
		Classification* &  \bf{4.33} & 12.84 & 16.05 & 8.88  \\
		Classification &  4.34 & \bf{12.67} & \bf{15.93} & \bf{8.71}  \\
		\hline		
	\end{tabular}
\end{table*}

\begin{table*}
	\vspace{1pt}
	\caption{Comparison between classification and regression on the KITTI validation dataset.}
	\label{table_KITTI_class_model}
	\renewcommand\arraystretch{1.0}	
	\centering	
	\begin{tabular}{l@{\hskip 0.8cm}c@{\hskip 0.60cm}c@{\hskip 0.60cm}c@{\hskip 0.60cm}c}
		\hline
		{Method} & {SILog} & {sqRel} & {absRel} & { irmse} \\ \hline
		\hline
		Regression*  & 14.55 & 2.26 & 9.33 & 11.45  \\
		Regression  & 14.49 & 2.28 & 9.44 & 11.63  \\		
		\hline
		Classification* &  13.42 & \bf{1.92} & \bf{7.86} & 9.72  \\		
		Classification &  \bf{13.34} & 1.95 & 8.01 & \bf{9.58}  \\		
		\hline		
	\end{tabular}
\end{table*}

\subsection{Effect of attention mechanism }
In order to reveal the effectiveness of the channel-wise attention mechanism, we conduct an ablation study.
Results are shown in Tables~\ref{table_Scannet_attention_model} and~\ref{table_KITTI_attention_model}. 
Specially, each metric in Table~\ref{table_Scannet_attention_model} is multiplied by 100 for easy comparison. 
In these tables, ``DABC w/o attention'' represents a DABC model that ignores the attention vectors and directly sums the multi-scale features in each fusion block.
Compared with ``DABC w/o attention'', our DABC model makes a significant improvement on the ScanNet dataset and achieves comparable performance on the KITTI dataset.
\begin{figure*}
	\begin{center}
		\includegraphics[scale=.68]{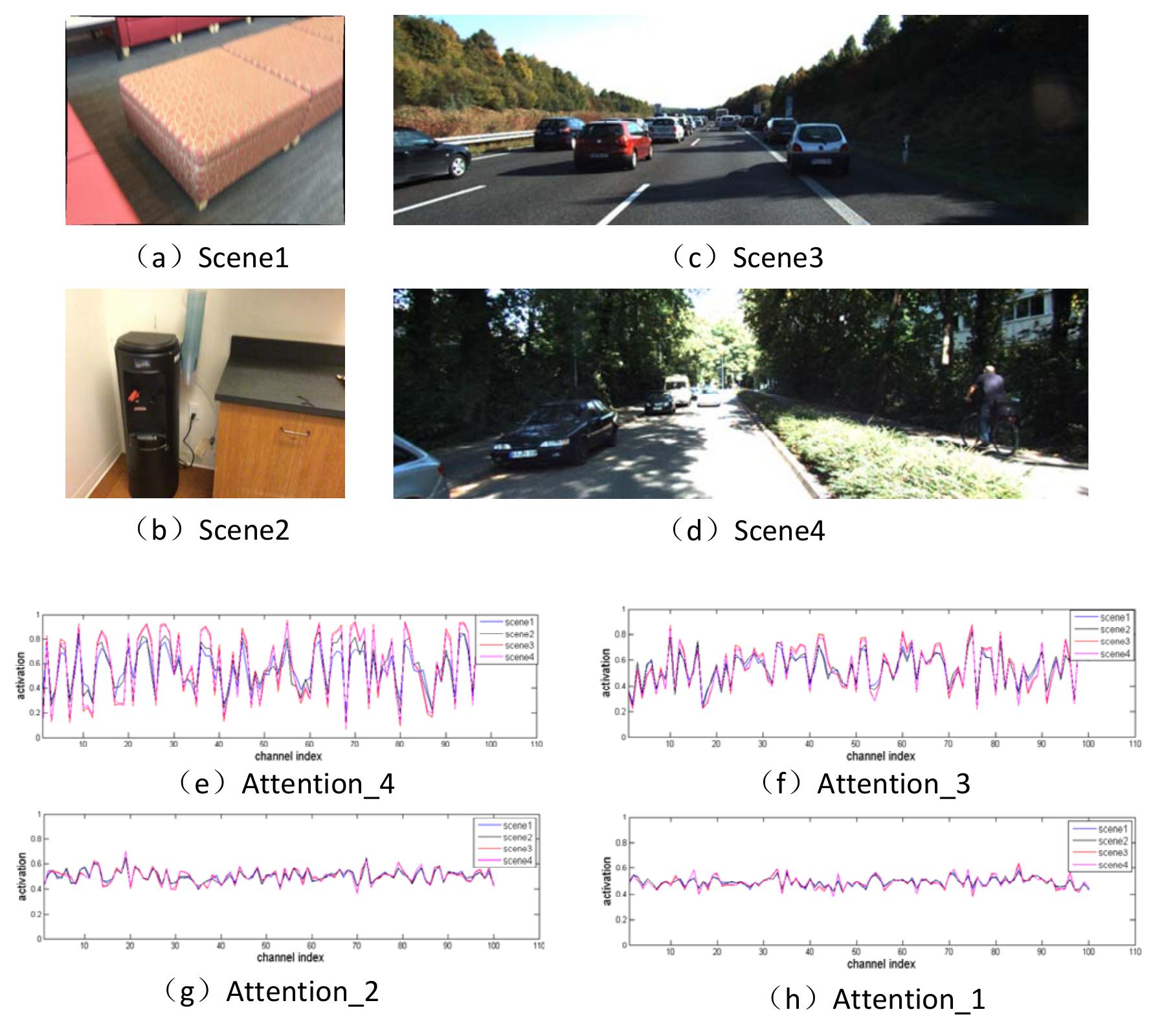}
	\end{center}
	\caption{Activations of different AFA modules. (a) and (b) are the color images from the ScanNet, (c) and (d) are from the KITTI. (e)$\sim$(h) are the activations of AFA modules from the high-level to the low-level in sequence. } 
	\label{fig_attention}
\end{figure*}

\begin{table*}
	\vspace{1pt}
	\caption{ Evaluation of the attention mechanism on the ScanNet validation dataset.}
	\label{table_Scannet_attention_model}
	\renewcommand\arraystretch{1.0}	
	\centering	
	\begin{tabular}{l@{\hskip 0.8cm}c@{\hskip 0.60cm}c@{\hskip 0.60cm}c@{\hskip 0.60cm}c}
		\hline
		{Method} & {sqRel} & {absRel} & { irmse} &{imae}\\ \hline
		\hline
		DABC w/o attention  & 4.93 & 13.50 & 16.37 & 9.073  \\
		DABC &  \bf{4.34} & \bf{12.67} & \bf{15.93} & \bf{8.71}  \\
		\hline		
	\end{tabular}
\end{table*}

\begin{table*}
	\vspace{1pt}
	\caption{ Evaluation for attention mechanism on the KITTI validation dataset.}
	\label{table_KITTI_attention_model}
	\renewcommand\arraystretch{1.0}	
	\centering	
	\begin{tabular}{l@{\hskip 0.8cm}c@{\hskip 0.60cm}c@{\hskip 0.60cm}c@{\hskip 0.60cm}c}
		\hline
		{Method} & {SILog} & {sqRel} & {absRel} & { irmse} \\ \hline
		\hline
		DABC w/o attention  & \bf{13.33} & 2.00 & \bf{8.00} & \bf{9.58}  \\
		DABC &  13.34 & \bf{1.95} & 8.01 & \bf{9.58}  \\		
		\hline		
	\end{tabular}
\end{table*}

Further, in order to visualize the activations, we choose four images from the KITTI and ScanNet datasets, and draw the activations of four AFA blocks, as shown in Figure~\ref{fig_attention}.  
For clarity, only one hundred activations of each block are visualized.
We make three observations about the attention mechanism in robust depth prediction.
First, we find that the activations of four scenes are more different in the high-level block than in the low-level one, which suggests that the values of each channel in the high-level features are scene-specific.
Second, as per Figure~\ref{fig_attention}(e), we observe a significant difference between indoor and outdoor activations. 
It indicates that the attention mechanism can give different scenes with different activations based on the characteristics and the layout of the scene.
Third, by comparing the activations of scene1 and scene2 in Figure~\ref{fig_attention}(e), we can also observe a non-negligible difference of activations between the two indoor scenes, which suggests that the attention mechanism still has a strong discriminating ability when processing the scenes from the same dataset.  

Therefore, we believe that the channel-wise attention mechanism plays a vital role in choosing discriminative features for diverse scenes and improving the performance in robust depth prediction.

\section{Conclusion}

In this paper, we study the task of robust depth prediction that requires a model suitable for both indoor and outdoor scenes with a single parameter set. 
Unlike conventional depth prediction tasks, robust depth prediction task needs the model to extract more discriminative features for diverse scenes and to tackle the large differences in depth ranges between indoor and outdoor scenes.  
To this end, we proposed a deep attention-based classification network to learn a universal RGB-to-depth mapping which is suitable for both indoor and outdoor scenes, where a channel-wise attention mechanism is employed to update the features according to the importance of each channel.
Experimental results on both indoor and outdoor datasets demonstrate the effectiveness of our method. 
Specifically, we won the 2nd place in the single image depth prediction entry of ROB 2018, in conjunction with CVPR 2018.
In the future, we plan to extend our method to other dense prediction tasks.

\bibliographystyle{splncs}
\bibliography{egbib}

\end{document}